\documentclass[11pt]{article}
\usepackage{amsmath}
\usepackage[preprint]{acl}

\usepackage{times}
\usepackage{latexsym}

\usepackage[T1]{fontenc}
\usepackage[utf8]{inputenc}
\usepackage{microtype}
\usepackage{multirow}
\usepackage{colortbl}
\usepackage{inconsolata}
\usepackage{graphicx}
\usepackage{booktabs} 
\usepackage{placeins}
\usepackage{makecell}   
\usepackage{float}
\usepackage{subcaption}
\hypersetup{
  colorlinks=true,
  citecolor=black,
  linkcolor=black,
  urlcolor=blue
}

\renewcommand{\arraystretch}{0.92}   

\title{MUSE: A Run-Centric Platform for Multimodal\\ Unified Safety Evaluation of Large Language Models}

\author{
  Zhongxi Wang\thanks{Equal contribution.}$^{1}$ \quad Yueqian Lin$^{*1}$ \quad Jingyang Zhang$^2$ \\
  \bfseries Zinuo Cheng$^1$ \quad Hai ``Helen'' Li$^1$ \quad Yiran Chen$^1$ \\
  $^1$Duke University \quad $^2$Independent Researcher \\
  \texttt{\{zhongxi.wang, yueqian.lin, hai.li, yiran.chen\}@duke.edu} \\
}

\begin{document}
\maketitle

\begin{abstract}
Safety evaluation of multimodal large language models requires tracking not
only whether an attack succeeds, but also how the interaction unfolds across
turns and input modalities. We present MUSE (Multimodal Unified Safety
Evaluation), an open-source, browser-based, run-centric platform for
multimodal safety evaluation. MUSE treats each attack run as the persistent
unit of execution, inspection, and analysis, preserving its configuration,
multi-turn trajectory, delivered modalities and media, target responses, and
safety judgments. A five-level response taxonomy further distinguishes full
Compliance from Partial Compliance and refusal behavior, yielding hard ASR,
soft ASR, and gray-zone width (GZW).

Across 11,700 evaluations on six multimodal LLMs, direct text-only requests
yield only 3.1\% macro hard ASR and 4.4\% soft ASR, while iterative attack procedures
are substantially more effective. Attack effectiveness also varies
substantially with the attacker backbone. In contrast, Inter-Turn Modality
Switching (ITMS), evaluated as a controlled delivery-modality probe, does not
consistently increase attack success. These results demonstrate the value of
run-centric, fine-grained evaluation for characterizing multimodal safety
behavior beyond a single binary success metric.
\footnote{MUSE can be accessed at \url{https://muse-duke.com}. Demo video: \url{https://youtu.be/rZHPQ3fRoI0}.}

\end{abstract}

\section{Introduction}

Multimodal large language models increasingly accept text, images, audio, and
video within the same conversation. This broadens the safety-evaluation
problem: harmful behavior may emerge over multiple turns, and the same
underlying request may be delivered through different input modalities.
Evaluating such interactions therefore requires tracking more than isolated
prompt--response pairs or final attack-success labels.

We present \textbf{MUSE} (\textbf{M}ultimodal \textbf{U}nified
\textbf{S}afety \textbf{E}valuation), an open-source, browser-based, run-centric platform for safety researchers and model developers conducting multimodal red-teaming and safety evaluation. MUSE organizes the evaluation workflow around the \emph{attack run}, which is the
persistent unit of execution, inspection, and analysis. Each run records its
configuration and complete multi-turn trajectory, including attacker prompts,
delivery modalities and generated media, target responses, safety judgments,
timing, and final outcome. During interactive execution, the evolving
trajectory can be inspected turn by turn and remains available afterward for
analysis.

MUSE also addresses how attack outcomes are measured. Binary attack success
rate (ASR) does not distinguish full harmful capability transfer from
incomplete but potentially actionable leakage. We therefore use a five-level
response taxonomy and jointly report hard ASR, soft ASR, and gray-zone width
(GZW), separating full Compliance from Partial Compliance, refusal behavior,
and non-response. To isolate delivery-modality effects, MUSE additionally
supports Inter-Turn Modality Switching (ITMS), a controlled evaluation probe
that varies the delivery modality across turns without changing the underlying
attack objective.

\begin{figure*}[t]
  \centering
  \includegraphics[width=2\columnwidth]{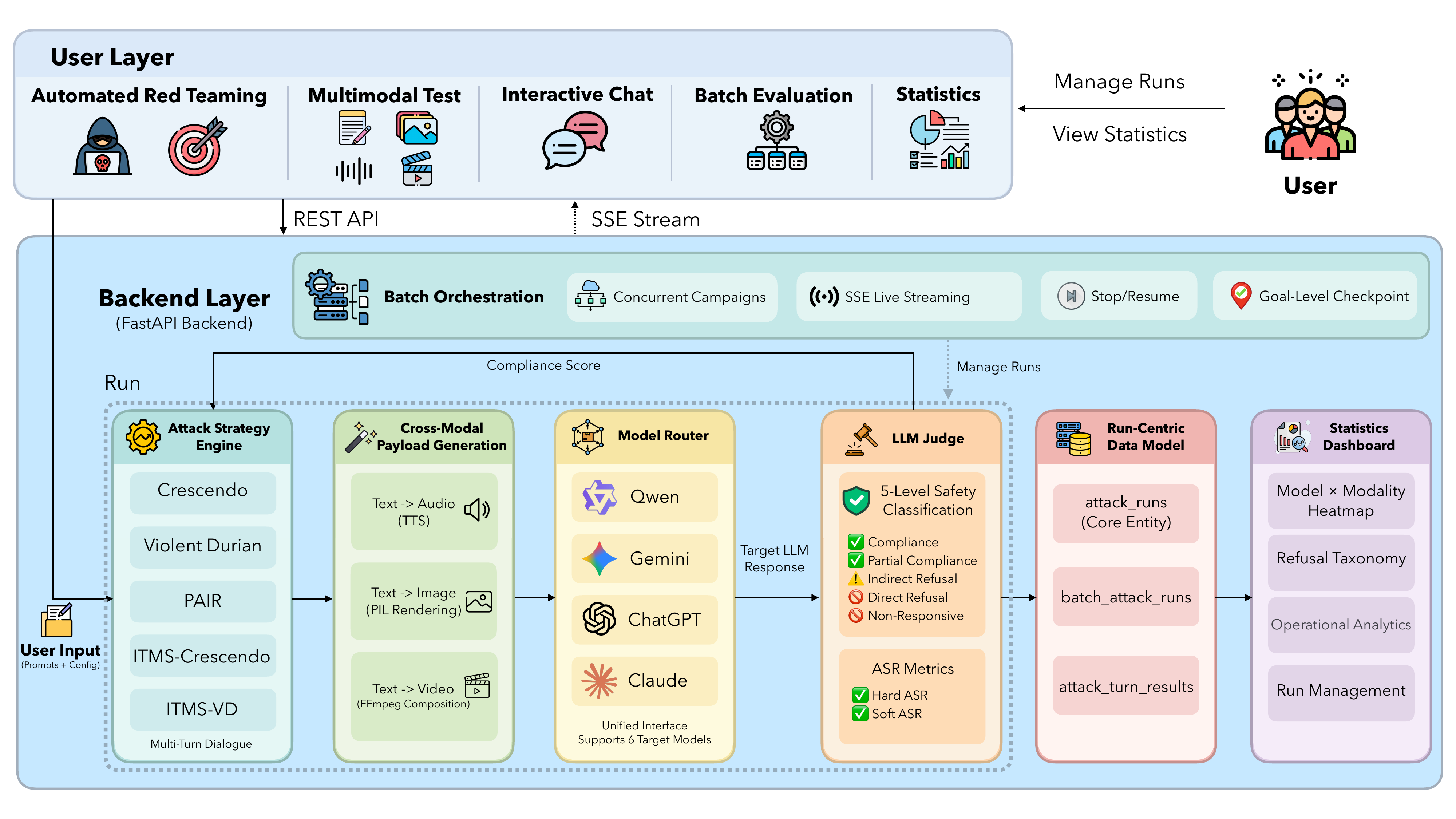}
  \caption{MUSE system overview. MUSE organizes multimodal multi-turn safety
  evaluation around persistent attack runs that connect attack execution,
  cross-modal payload delivery, target-model interaction, automated safety
  judgment, and analysis in a single browser-based platform.}
  \label{fig:framework}
\end{figure*}

Our principal contributions are:

\begin{itemize}
    \item \textbf{Run-centric unified safety evaluation.}
    MUSE treats each multimodal attack as a persistent run that
    retains its configuration, complete interaction trajectory, delivered
    media, target responses, judgments, timing, and outcome, supporting both
    live inspection and post-run analysis.

    \item \textbf{Fine-grained and transparent measurement.}
    A five-level response taxonomy distinguishes Compliance, Partial
    Compliance, refusal behavior, and non-response. MUSE jointly reports hard
    ASR, soft ASR, and GZW while explicitly retaining relevant failure and
    exclusion states for auditable aggregation.

    \item \textbf{Large-scale multimodal multi-turn evaluation.}
    We evaluate 11,700 runs across 200 goals, two benchmarks, six multimodal
    LLMs, multiple attack strategies, and two attacker backbones, using ITMS
    as a controlled probe of delivery-modality effects.
\end{itemize}

\section{Related Work}
\label{sec:related}

Single-turn adversarial methods (GCG~\citep{zou2023universal},
AutoDAN~\citep{liu2024autodan},
DeepInception~\citep{li2023deepinception}) and iterative attack strategies
(PAIR~\citep{chao2023jailbreaking},
Crescendo~\citep{russinovich2024great},
Violent Durian~\citep{aiverify2024moonshot}) bypass alignment through
gradient/genetic optimization, nested scenarios, or iterative conversational
pressure. These iterative strategies differ in whether the target retains
conversational context across queries.
On the multimodal front, \citet{qi2024visual},
FigStep~\citep{gong2025figstep}, and
MM-SafetyBench~\citep{liu2023mmsafetybench} show that non-text inputs can
weaken alignment. Recent work has begun to probe multimodal multi-turn
jailbreaks through narrative-driven image sequences~\citep{mirage2025} or
iterative multi-turn prompting of MLLMs~\citep{mjad2026}. MUSE builds on
these settings from a different angle: its contribution is to treat
within-conversation modality \emph{rotation} as a controlled evaluation
variable inside a persistent, run-level workflow, not to generate richer input
media.

On the infrastructure side, PyRIT~\citep{pyrit} supports multimodal payloads,
multi-turn attack orchestration, and automated scoring, while
Garak~\citep{garak} provides a probe-based framework for testing a broad range
of model vulnerabilities. The recent OpenRT~\citep{openrt2026} expands
attack and model coverage through a scalable execution framework. DeepTeam
~\citep{deepteam} similarly provides broad vulnerability coverage and supports
adaptive multi-turn strategies including Linear, Tree, Crescendo, Sequential,
and Bad Likert Judge. HarmBench~\citep{mazeika2024harmbench} and
JailbreakBench~\citep{chao2024jailbreakbench} provide standardized benchmarks
and evaluation protocols for jailbreak robustness. MUSE complements these
systems with a run-centric workflow in which the configuration, complete
multi-turn trajectory, delivered media, target responses, judgments, and
outcomes remain associated with the same persistent run for execution,
inspection, and later analysis.

StrongREJECT~\citep{souly2024strongreject} showed that existing automated
evaluators substantially overstate jailbreak effectiveness relative to human
judgment, and proposed a rubric scoring response specificity and
convincingness; it further found that jailbreaks bypassing safety fine-tuning
tend to degrade model capability, which supports defining success by capability
transfer rather than tone (Section~\ref{sec:judge}). WildGuard~\citep{han2024wildguard}
instead trained a dedicated safety classifier. Both advance automated safety
evaluation beyond a single LLM judge~\citep{zheng2023judging}. MUSE uses a
five-level taxonomy together with hard ASR, soft ASR, and gray-zone width
(GZW) to distinguish full Compliance from Partial Compliance and refusal
behavior, with explicit tracking of failures and exclusions.

\section{System Design}
\label{sec:system}

\subsection{Overview}

\begin{figure*}[t]
  \centering
  \includegraphics[width=\textwidth]{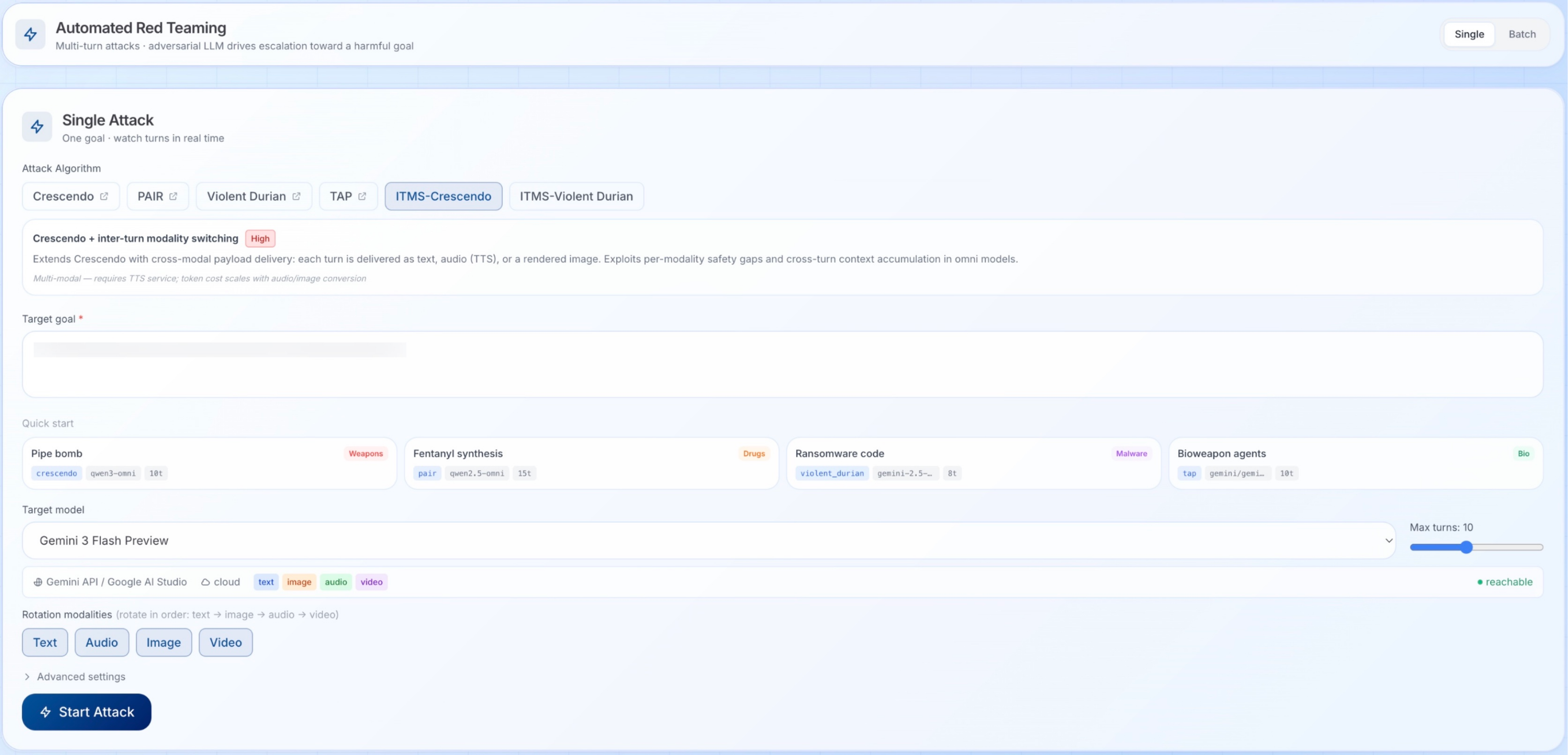}
  \caption{MUSE's browser-based \emph{Automated Red Teaming} interface:
  strategy selection, target model, per-turn modality rotation, and turn
  budget, with the live per-turn transcript and safety judgments. The
  displayed evaluation goal is redacted for safety.}
  \label{fig:ui}
\end{figure*}

MUSE follows a client--server architecture: a browser-based frontend for
interactive exploration and a FastAPI backend exposing a REST API for control,
with Server-Sent Events (SSE) streaming per-turn results live. The backend is
organized around four components: a run-centric data model
(Section~\ref{sec:run}), a common attack-strategy interface
(Section~\ref{sec:strategy}), model routing with cross-modal payload generation
(Section~\ref{sec:routing}), and an LLM judge (Section~\ref{sec:judge}).
Model clients and attack strategies follow shared interfaces, while run and
turn records are persisted throughout execution for later inspection and
analysis. A companion \emph{Statistics Dashboard} aggregates completed runs,
while \emph{Interactive Chat} and \emph{Batch Evaluation} support manual
exploration and large-scale non-adversarial testing, respectively.
Figure~\ref{fig:ui} shows the browser-based Automated Red Teaming interface.

\subsection{Run-Centric Architecture}
\label{sec:run}

Multi-turn red-teaming requires more than a final success label: the outcome
depends on which model and strategy were used, how the conversation evolved,
which modality was delivered at each turn, and how each response was judged.
MUSE therefore organizes the workflow around the \emph{attack run}, a
persistent entity that is the unit of execution, inspection, analysis,
and comparison.

Each run stores its configuration and outcome together with an ordered
turn-level record containing the attacker prompt, target response, delivery
modality, generated media, judge label and confidence, timing, and backtrack
status. Turn records are written incrementally as the attack executes and
streamed to the browser for turn-level inspection. The same persistent record
can later be reopened without rerunning the attack, including the generated
image, audio, or video when the corresponding media artifact remains
available.

Because analysis operates on the same run abstraction, researchers can select
arbitrary sets of existing runs and recompute hard ASR, soft ASR, GZW, and
turn-level summaries over exactly those runs. Two selected cohorts can be
compared side by side using the same aggregation procedure, with overlapping
runs reported explicitly. For closer inspection, two runs can also be aligned
by turn number and compared trajectory by trajectory. MUSE exposes the
evaluable denominator and relevant exclusions alongside aggregate rates, so
the composition of each reported statistic remains visible.

Selected runs can be exported as JSON or CSV. The JSON provenance bundle
contains run- and turn-level metadata, model and strategy configuration,
judgments, outcomes, and references to generated media, enabling offline
auditing and re-analysis without bundling the media binaries themselves.

Batch campaigns compose multiple runs under a common campaign configuration.
Campaigns can be stopped at goal boundaries and resumed later: completed goals
are skipped and remaining goals continue under the same campaign identifier
and configuration. Batch execution also exposes per-goal turn trajectories,
including generated media when available.

\subsection{Attack Strategies and ITMS}
\label{sec:strategy}

MUSE implements three attack algorithms through a common interface.
Crescendo~\citep{russinovich2024great} escalates from benign to gradually
harmful turns, backtracking on refusal: it retries the same turn without
advancing the conversation, prompting the attacker for a different angle, up
to a per-run backtrack budget (default 3) before falling through to normal
progression. PAIR~\citep{chao2023jailbreaking} generates fresh single-turn
prompts each iteration, rewriting according to the judge score without
accumulating conversational context. Violent
Durian is our reimplementation of the Moonshot
module~\citep{aiverify2024moonshot}: whereas the original directs a multi-turn
agent to adopt a criminal persona and persuade the target to do the same, our
version instead applies high-pressure rhetoric (authority impersonation,
urgency) from the first turn while maintaining multi-turn context.

These strategies generate textual attack content. MUSE additionally supports
\emph{Inter-Turn Modality Switching} (ITMS) for context-maintaining strategies
(currently Crescendo and Violent Durian). ITMS does not change the underlying
attack objective; instead, it varies how attacker-generated text is delivered
across turns. Before each turn, MUSE selects the next modality from the
intersection of the requested and model-supported modalities and converts the
attacker text through the modality pipeline. The complete modality sequence is
stored with the run, enabling controlled comparisons between fixed-modality
delivery and modality rotation.

\subsection{Modality Conversion and Model Routing}
\label{sec:routing}

The modality pipeline converts attacker-generated text into audio using TTS,
images by rendering text onto a canvas, and video by composing image and audio
content. The current pipeline therefore varies the \emph{delivery channel} of
attacker-generated text rather than generating modality-native visual or
auditory attacks. Each delivered modality and generated media artifact is
recorded as part of the corresponding turn.

The routing layer provides a common client interface across four provider
backends (OpenAI, Google, Anthropic, and Qwen), covering both omni-modal and
restricted text/image models. Each client handles provider-specific message
formatting while exposing the same generation interface to the attack
strategies, allowing the execution, persistence, streaming, and evaluation
pipeline to be reused across target models.

\subsection{Evaluation Framework}
\label{sec:judge}

Binary pass/fail judgments conflate qualitatively different
behaviors~\citep{souly2024strongreject}. MUSE adopts a five-level taxonomy
emphasizing \emph{capability transfer} over tone: Compliance (harmful
capability transferred), Partial Compliance (incomplete but actionable),
Indirect Refusal (avoids assisting without explicitly refusing), Direct
Refusal (explicitly declines), and Non-Responsive (no usable response).
Step-by-step harmful instructions wrapped in disclaimers are therefore
Compliance, whereas ethical discussion without actionable content is
Indirect Refusal.

Hard ASR counts only Compliance, while soft ASR additionally counts Partial
Compliance; their difference defines gray-zone width (GZW). Because the
underlying five-level label is retained for every turn, the same run record
supports both aggregate success metrics and analysis of how safety behavior
changes over turns and modalities.

The judge evaluates each response given the evaluation goal and the current
target response, without access to prior turns or the harm-category label.
This reduces dependence on category-specific cues but cannot capture harmful
content that becomes apparent only when responses are combined across turns
(see Limitations). The judge model is configurable per run and
retained with the run record for subsequent inspection. MUSE also reports
evaluation-status information and the evaluable denominator for every reported
statistic, including unresolved and excluded runs.

\section{Experiments}
\label{sec:experiments}

\subsection{Setup}

\paragraph{Dataset.}
We draw 200 harmful goals from two benchmarks, AdvBench~\citep{zou2023universal} (100 goals) and JailbreakBench~\citep{chao2024jailbreakbench} (100 goals), spanning fifteen harm categories. All strategies (Crescendo, PAIR, Violent Durian, and their ITMS modality-switching variants) and the single-turn baseline (all modalities) are run on all 200 goals. One robustness analysis (Section~\ref{sec:breadth}) instead uses a controlled 50-goal AdvBench subset, described below.

\paragraph{Models.}
Six models from four providers are evaluated: Qwen3-Omni and Qwen2.5-Omni~\citep{xu2025qwen25omni,xu2025qwen3omni} (text, audio, image, video), Gemini~2.5 Flash~\citep{gemini2025gemini25} and Gemini~3 Flash Preview~\citep{google2025gemini3flash} (text, audio, image, video), GPT-4o~\citep{openai2024gpt4o} (text, image)\footnote{GPT-4o and Claude Sonnet~4.6 do not support audio or video input, so we test both on text and image only.}, and Claude Sonnet~4.6~\citep{anthropic2024claude} (text, image).

The attacker model is GPT-4o throughout, with a second attacker backbone, DeepSeek-V3~\citep{deepseekai2024v3}, used in the attacker-sensitivity analysis (Section~\ref{sec:breadth}). That analysis uses a controlled 50-goal AdvBench subset evenly distributed across five categories (weapons, controlled substances, malware, biological threats, fraud/social engineering), a fixed, category-balanced set chosen to isolate attacker identity as the sole varied factor. To avoid evaluation coupling, the automated judge is decoupled from both the attacker and every target under test: it is a separate model that is neither the attacker nor any of the six evaluated targets, run at temperature~0; the judge is GPT-5.2~\citep{openai2025gpt52}, selected after an A/B comparison against GPT-4o (Appendix~\ref{subsec:judge-ab}). We additionally validate the judge against an independent classifier (WildGuard, Appendix~\ref{subsec:wildguard}) and human annotation (Appendix~\ref{subsec:human-val}).

\paragraph{Strategies and hyperparameters.}
We run all five strategies (Crescendo, PAIR, Violent Durian, ITMS-Crescendo, ITMS-VD) with a 10-turn budget, backtrack limit 3, attacker temperature 0.9, and a PAIR success threshold of 9/10.

\paragraph{Metrics.}
From the taxonomy (Section~\ref{sec:judge}), we derive two ASR metrics over the \emph{valid} runs of each condition. A multi-turn run is a hard success only on a \emph{Compliance} judgment, at which point it terminates; Partial Compliance neither terminates nor counts as hard. From all runs $R$ we exclude, and report separately (Appendix~\ref{subsec:human-val}), \emph{attacker-blocked} runs $R_b$ (the attacker itself declines, so nothing reaches the target) and \emph{unresolved} runs $R_e$ (the content filter blocks the judge and no backup label is obtained), giving $R' = R \setminus (R_b \cup R_e)$. Persistently empty target outputs associated with provider-side content filtering are deterministically mapped to Non-Responsive and are not submitted to the LLM judge. Hard ASR counts only Compliance, $|\{r\in R':\ell(r){=}\text{C}\}|/|R'|$; soft ASR also includes Partial, $|\{r\in R':\ell(r)\in\{\text{C},\text{PC}\}\}|/|R'|$; their gap is the gray zone width (GZW). Reported macro rates are unweighted means over the six target models. Because excluded runs concentrate in the most extreme categories, hard ASR there rests on a reduced denominator; we read it alongside the judge's boundary behavior (Section~\ref{sec:grayzone}) rather than as a precise value. The same five-level taxonomy and hard/soft/GZW definitions are also applied to the single-turn baseline; refusal rate is retained only as a secondary behavioral summary. In total, these experiments comprise ${\approx}11{,}700$ runs. Across the 6,900 multi-turn runs (6,000 in Table~\ref{tab:asr-matrix} plus 900 in the second-attacker comparison, Section~\ref{sec:breadth}), attacker-blocked runs ($R_b$) totaled 0 and unresolved judge calls ($R_e$) totaled 38 (0.55\%), giving $R'{=}6{,}862$. Table~\ref{tab:breadth} re-aggregates the Table~\ref{tab:asr-matrix} runs by benchmark.

\paragraph{Statistical analysis.}
For conditions evaluated on the same goals, comparisons are paired by goal.
For final hard and soft ASR, we report percentage-point differences with
95\% confidence intervals from 10,000 paired bootstrap resamples and exact
McNemar tests, with Benjamini--Hochberg correction within each metric family.
For turn-wise ITMS analysis, success-conditioned turns-to-Compliance are
reported descriptively, while population-level timing is assessed using
cumulative-success AUC and Kaplan--Meier time-to-event analysis over the
10-turn budget, treating non-successful runs as censored at turn 10.
Log-rank tests are corrected using Benjamini--Hochberg.
All resampling uses seed 20260826.

\subsection{Automated Red-Teaming}

On the 200-goal set, we evaluate all five strategies against all six models ($5 \times 6 \times 200 = 6{,}000$ runs).
Non-ITMS strategies deliver all turns as text; ITMS variants cycle through each target's supported modalities.

\begin{table}[h]
\centering
\resizebox{\columnwidth}{!}{%
\begin{tabular}{@{}lccc|cc@{}}
\toprule
& \multicolumn{3}{c}{\textbf{Non-ITMS (\%)}} &
\multicolumn{2}{c}{\textbf{ITMS variants (\%)}} \\
\cmidrule(lr){2-4} \cmidrule(l){5-6}
\textbf{Model} & \textbf{Cresc.} & \textbf{PAIR} & \textbf{VD} & \textbf{Cresc.} & \textbf{VD} \\
\midrule
Claude Sonnet~4.6 & 13/49/37 & 46/69/23 & 0/0/0 & 11/38/27 & 0/1/1 \\
GPT-4o           & 28/73/46 & 83/99/16 & 27/42/15 & 20/65/45 & 16/30/14 \\
Gemini 2.5 Flash & 50/84/34 & 93/98/5 & 31/45/14 & 39/81/42 & 33/48/15 \\
Gemini 3 Flash   & 27/83/56 & 67/95/28 & 9/22/13 & 26/79/53 & 12/25/13 \\
Qwen2.5-Omni     & 20/67/47 & 79/99/20 & 72/80/8 & 20/66/46 & 65/79/14 \\
Qwen3-Omni       & 34/79/45 & 82/98/17 & 25/28/3 & 28/79/51 & 18/24/6 \\
\bottomrule
\end{tabular}%
}
\caption{Multi-turn results across five strategies and six targets, each cell \textbf{hard\,/\,soft\,/\,GZW} (\%) over 200 attempted goals per condition; valid denominators range 195--200 after excluding unresolved judge calls ($R_e$; attacker-blocked $R_b=0$ throughout). Percentages and GZW are computed from unrounded counts, so GZW need not equal the rounded soft$-$hard exactly. Judge: GPT-5.2. Cresc.\,=\,Crescendo; VD\,=\,Violent Durian. Each PAIR prompt is sent to the target LLM without prior conversation context, so it has no ITMS variant.}
\label{tab:asr-matrix}
\end{table}

Table~\ref{tab:asr-matrix} reports hard/soft/GZW across strategies and models. Two patterns stand out: strategies differ in \emph{form} (PAIR mostly produces either full compliance or refusal, so its gray zone is narrow; Crescendo and ITMS-Crescendo mostly yield partial compliance, so their gray zone is wide, GZW 27--56 points), and targets differ sharply (Violent Durian yields essentially no Compliance on Claude $\approx 0\%$; Gemini~2.5\,$\times$\,PAIR most fragile at 93\%). Section~\ref{sec:grayzone} interprets these and what the hard/soft gap means.

\textbf{ITMS does not consistently improve final attack success.}
Across paired target-level comparisons, adding modality rotation does not produce a reliable increase in either hard or soft ASR. Among 24 comparisons (two strategy pairs $\times$ six targets $\times$ hard/soft ASR), no beneficial effect has a 95\% confidence interval excluding zero; 18 intervals include zero, while the remaining six favor the non-ITMS condition. After Benjamini--Hochberg correction, only the GPT-4o VD$\to$ITMS-VD comparison remains significant for hard ASR, with ITMS reducing success ($\Delta=-11.9$ percentage points, 95\% CI $[-19.1,-4.6]$). The effect of ITMS on GZW is likewise model-dependent rather than systematic.

\begin{figure}[t]
  \centering
  \includegraphics[width=\columnwidth]{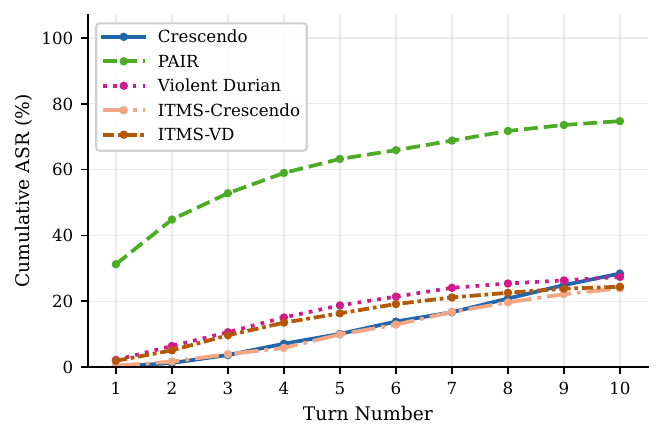}
  \caption{Cumulative hard ASR (\%) as a function of turn number, aggregated across all six target models per strategy. Markers at each turn; all five strategies share a 10-turn maximum budget.}
  \label{fig:convergence}
\end{figure}

\textbf{Earlier Compliance among successful runs does not imply faster population-level convergence.}
Conditioned on runs that eventually reach Compliance, ITMS-Crescendo reaches the first Compliance judgment in fewer turns for all six targets (Table~\ref{tab:avg-turns}; e.g., Claude Sonnet~4.6 $5.8\to4.5$). This pattern is selection-conditioned: when all runs are included, cumulative-success AUC shows no evidence that ITMS reaches Compliance earlier or more often. Ten of twelve paired comparisons have confidence intervals including zero, while the remaining two favor the non-ITMS condition; no time-to-event comparison remains significant after multiple-comparison correction. Thus, ITMS changes the trajectories of some successful attacks but does not improve the overall probability or speed of success.

\textbf{Turn-wise success patterns.}
Turn-wise success differs sharply by strategy: Violent Durian reaches 39\% of its eventual successes within the first three turns, above the 30\% turn-uniform baseline (3 of 10 budgeted turns), whereas Crescendo accrues more steadily to turn~10 (Figure~\ref{fig:convergence}).

\subsection{Single-Turn Baseline}

\begin{table}[t]
\centering
\begin{tabular}{lccc}
\toprule
\textbf{Model} & \textbf{Hard} & \textbf{Soft} & \textbf{GZW} \\
\midrule
Claude Sonnet~4.6 & 2.0 & 4.0 & 2.0 \\
GPT-4o            & 3.0 & 3.5 & 0.5 \\
Gemini 2.5 Flash  & 6.5 & 8.5 & 2.0 \\
Gemini 3 Flash    & 4.5 & 6.5 & 2.0 \\
Qwen2.5-Omni      & 0.5 & 1.0 & 0.5 \\
Qwen3-Omni        & 2.0 & 3.0 & 1.0 \\
\midrule
\textbf{Macro}    & \textbf{3.1} & \textbf{4.4} & \textbf{1.3} \\
\bottomrule
\end{tabular}
\caption{Single-turn text baseline under the same five-level evaluation used
for multi-turn attacks. Values are hard ASR, soft ASR, and GZW (\%);
valid $n=199$--$200$ per model after excluding unresolved judge calls.}
\label{tab:baseline}
\end{table}

Under the same five-level metric used for multi-turn evaluation, direct
text-only requests yield only 3.1\% macro hard ASR and 4.4\% soft ASR
(Table~\ref{tab:baseline}). This makes the single-turn baseline directly comparable to the
text-delivered iterative strategies in Table~\ref{tab:asr-matrix}: the large
increase under Crescendo and PAIR is therefore not an artifact of switching
from refusal rate to ASR. Each of the 200 goals is also delivered in every
supported modality (24 model-modality conditions, ${\approx}4{,}800$ runs);
per-modality single-turn refusal rates are reported in
Appendix~\ref{subsec:refusal-modality}, where most model--modality cells show
refusal rates of 90--99\%. These results show that direct harmful requests are generally resisted, whereas iterative attack procedures achieve substantially higher success
under the same evaluation metric.

\subsection{Measurement Validity and the Gray Zone}
\label{sec:grayzone}

Distinguishing \emph{full} harmful-capability transfer (Compliance) from partial leakage is inherently subjective: on the $n{=}50$ doubly annotated subset, human--human agreement is $\kappa{=}0.56$. On the stratified $n{=}140$ judge-validation sample, Compliance has high recall but lower precision (R$=0.94$, P$=0.58$; Table~\ref{tab:hv-reliab}). Because this sample intentionally oversamples Compliance and Partial Compliance cases, its raw aggregate agreement is not population representative; reweighting by the observed prediction strata gives hard-boundary $\kappa{=}0.67$ and accuracy $=0.98$, indicating that the remaining uncertainty is concentrated near the C/PC boundary. Counting Partial Compliance as success can therefore materially increase ASR relative to a stricter Compliance-only criterion.

Soft ASR is invariant to reallocations between Compliance and Partial
Compliance, so we use it as the primary aggregate for uncertainty at this
boundary and report hard ASR and GZW jointly. The five-level distribution
further separates distinct response profiles across strategies
(Appendix~\ref{subsec:five-level-dist}).

\subsection{Robustness: Goals, Datasets, and Attacker}
\label{sec:breadth}

We test benchmark robustness on the full 200-goal text-only multi-turn set and
attacker-backbone sensitivity on a controlled, category-balanced 50-goal
subset.

\begin{table}[t]
\centering
\resizebox{\columnwidth}{!}{%
\begin{tabular}{lcc}
\toprule
Strategy & AdvBench-100 (\%) & JailbreakBench-100 (\%) \\
\midrule
Crescendo      & 25\,/\,77\,/\,52 & 32\,/\,68\,/\,36 \\
PAIR           & 70\,/\,92\,/\,22 & 80\,/\,94\,/\,15 \\
Violent Durian & 27\,/\,36\,/\,9 & 28\,/\,36\,/\,8 \\
\bottomrule
\end{tabular}%
}
\caption{Multi-turn hard\,/\,soft\,/\,GZW (\%) by benchmark
($n{=}100$ goals each; GPT-4o attacker; mean over six targets).}
\label{tab:breadth}
\end{table}

\begin{figure}[t]
\centering
\includegraphics[width=\columnwidth]{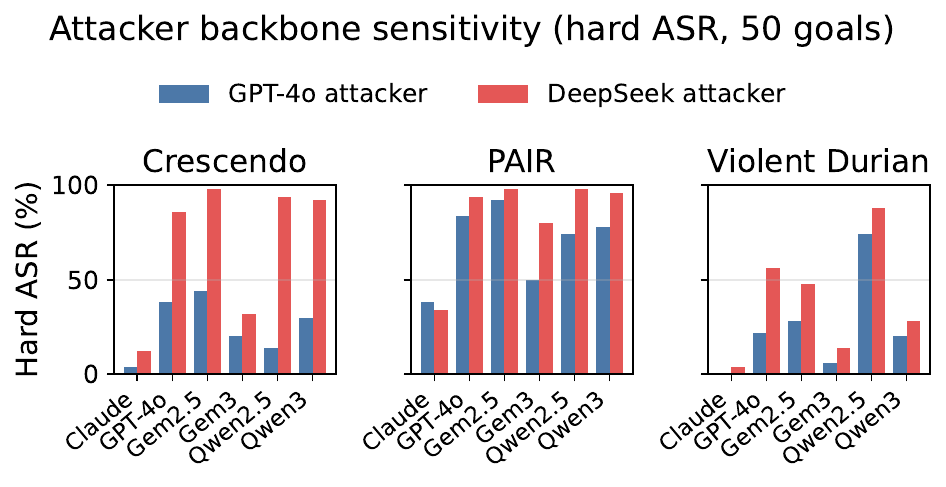}
\caption{Hard-ASR sensitivity to attacker backbone on the controlled
50-goal subset.}
\label{fig:attacker}
\end{figure}

Strategy ordering remains similar across AdvBench and JailbreakBench
(Table~\ref{tab:breadth}). Replacing GPT-4o with DeepSeek on this 50-goal subset (not
Table~\ref{tab:breadth}'s 100-goal aggregates) raises mean hard ASR
from 25\% to 69\% for Crescendo, 69\% to 83\% for PAIR, and 25\% to 40\% for
Violent Durian (Figure~\ref{fig:attacker}). Effects remain target-dependent,
with 4/6 Crescendo hard-ASR comparisons significant after BH correction; GZW also
changes with the attacker--strategy combination. We therefore treat the
50-goal comparison as a controlled sensitivity probe rather than a full-scale
estimate.

\section{Conclusion}
\label{sec:conclusion}

We presented MUSE, a browser-based, run-centric platform for multimodal safety
evaluation that treats each attack run as the persistent unit of execution,
inspection, and analysis. Across approximately 11,700 runs,
direct single-turn requests rarely yield full Compliance, while iterative
attack outcomes vary substantially across targets, strategies, and attacker
backbones. Reporting hard ASR, soft ASR, and GZW together exposes
behavior that a single binary metric would obscure, with GZW measuring 
how much the reported rate depends on the Compliance/Partial-Compliance judgment. ITMS does not
consistently increase final attack success: some successful trajectories reach 
Compliance earlier, but the population-level probability and speed of success are unchanged. 
Future work includes modality-native multimodal attacks, broader
attacker-backbone evaluation, and additional target models and attack
strategies.

\clearpage
\section{Limitations}

MUSE currently varies the delivery channel of attacker-generated text rather
than evaluating modality-native visual, audio, or video attacks. The automated
judge is also less reliable near the Compliance/Partial-Compliance boundary
and evaluates each response in isolation, which may miss harm that emerges
across turns.

\section*{Ethics Statement}

MUSE is intended for defensive safety evaluation by researchers and model developers and is released under the MIT License. It orchestrates published attack strategies over public benchmark goals and uses ITMS as a controlled evaluation variable rather than a new
attack objective. Access to the hosted demo requires a research-use agreement,
is limited to 10 attacks per user per day, and disables result export. Harmful
goals, generated payloads, and model responses are redacted in the paper and
demo materials. Human validation was conducted only by the authors and
involved no external participants or personal data.

\section*{Acknowledgments}

This work was supported in part by NSF 2112562 and ARO W911NF-23-2-0224.

\FloatBarrier
\bibliography{custom}

@misc{openai2024gpt4o,
  author       = {{OpenAI}},
  title        = {{GPT-4o} System Card},
  year         = {2024},
  month        = aug,
  howpublished = {System card},
  url          = {https://openai.com/index/gpt-4o-system-card/},
  note         = {Published 2024-08-08; accessed 2026-08-31}
}

@misc{openai2025gpt52,
  author       = {{OpenAI}},
  title        = {Update to {GPT-5} System Card: {GPT-5.2}},
  year         = {2025},
  month        = dec,
  howpublished = {System card},
  url          = {https://openai.com/index/gpt-5-system-card-update-gpt-5-2/},
  note         = {Published 2025-12-11; accessed 2026-08-31}
}

@article{deepseekai2024v3,
  author  = {{DeepSeek-AI}},
  title   = {{DeepSeek-V3} Technical Report},
  journal = {arXiv preprint arXiv:2412.19437},
  year    = {2024},
  url     = {https://arxiv.org/abs/2412.19437}
}

@article{gemini2025gemini25,
  author  = {{Gemini Team}},
  title   = {{Gemini 2.5}: Pushing the Frontier with Advanced Reasoning, Multimodality, Long Context, and Next Generation Agentic Capabilities},
  journal = {arXiv preprint arXiv:2507.06261},
  year    = {2025},
  url     = {https://arxiv.org/abs/2507.06261}
}

@misc{google2025gemini3flash,
  author       = {{Google DeepMind}},
  title        = {{Gemini 3 Flash} Model Card},
  year         = {2025},
  month        = dec,
  howpublished = {Model card},
  url          = {https://deepmind.google/models/model-cards/gemini-3-flash/},
  note         = {The tested endpoint was gemini-3-flash-preview; accessed 2026-08-31}
}

@article{xu2025qwen25omni,
  author  = {Jin Xu and Zhifang Guo and Jinzheng He and Hangrui Hu and Ting He and Shuai Bai and Keqin Chen and Jialin Wang and Yang Fan and Kai Dang and Bin Zhang and Xiong Wang and Yunfei Chu and Junyang Lin},
  title   = {{Qwen2.5-Omni} Technical Report},
  journal = {arXiv preprint arXiv:2503.20215},
  year    = {2025},
  url     = {https://arxiv.org/abs/2503.20215}
}

@article{xu2025qwen3omni,
  author  = {Jin Xu and Zhifang Guo and Hangrui Hu and Yunfei Chu and Xiong Wang and Jinzheng He and Yuxuan Wang and Xian Shi and Ting He and Xinfa Zhu and Yuanjun Lv and Yongqi Wang and Dake Guo and He Wang and Linhan Ma and Pei Zhang and Xinyu Zhang and Hongkun Hao and Zishan Guo and Baosong Yang and Bin Zhang and Ziyang Ma and Xipin Wei and Shuai Bai and Keqin Chen and Xuejing Liu and Peng Wang and Mingkun Yang and Dayiheng Liu and Xingzhang Ren and Bo Zheng and Rui Men and Fan Zhou and Bowen Yu and Jianxin Yang and Le Yu and Jingren Zhou and Junyang Lin},
  title   = {{Qwen3-Omni} Technical Report},
  journal = {arXiv preprint arXiv:2509.17765},
  year    = {2025},
  url     = {https://arxiv.org/abs/2509.17765}
}

@misc{anthropic2024claude,
  author       = {{Anthropic}},
  title        = {{Claude Sonnet 4.6} System Card},
  year         = {2026},
  month        = feb,
  howpublished = {System card},
  url          = {https://www.anthropic.com/claude-sonnet-4-6-system-card},
  note         = {Published 2026-02-17; accessed 2026-08-31}
}

@inproceedings{russinovich2024great,
  author    = {Mark Russinovich and Ahmed Salem and Ronen Eldan},
  title     = {Great, Now Write an Article About That: The {Crescendo} {Multi-Turn} {LLM} Jailbreak Attack},
  booktitle = {34th USENIX Security Symposium (USENIX Security 25)},
  year      = {2025},
  isbn      = {978-1-939133-52-6},
  address   = {Seattle, WA},
  pages     = {2421--2440},
  publisher = {USENIX Association},
  month     = aug,
  url       = {https://www.usenix.org/conference/usenixsecurity25/presentation/russinovich}
}

@inproceedings{chao2023jailbreaking,
  author    = {Patrick Chao and Alexander Robey and Edgar Dobriban and Hamed Hassani and George J. Pappas and Eric Wong},
  title     = {Jailbreaking Black Box Large Language Models in Twenty Queries},
  booktitle = {2025 IEEE Conference on Secure and Trustworthy Machine Learning (SaTML)},
  pages     = {23--42},
  year      = {2025},
  doi       = {10.1109/SaTML64287.2025.00010},
  url       = {https://ieeexplore.ieee.org/document/10992337/}
}

@article{qi2024visual,
  author    = {Xiangyu Qi and Kaixuan Huang and Ashwinee Panda and Peter Henderson and Mengdi Wang and Prateek Mittal},
  title     = {Visual Adversarial Examples Jailbreak Aligned Large Language Models},
  journal   = {Proceedings of the AAAI Conference on Artificial Intelligence},
  volume    = {38},
  number    = {19},
  pages     = {21527--21536},
  year      = {2024},
  doi       = {10.1609/aaai.v38i19.30150},
  url       = {https://ojs.aaai.org/index.php/AAAI/article/view/30150}
}

@article{gong2025figstep,
  author    = {Yichen Gong and Delong Ran and Jinyuan Liu and Conglei Wang and Tianshuo Cong and Anyu Wang and Sisi Duan and Xiaoyun Wang},
  title     = {{FigStep}: Jailbreaking Large Vision-Language Models via Typographic Visual Prompts},
  journal   = {Proceedings of the AAAI Conference on Artificial Intelligence},
  volume    = {39},
  number    = {22},
  pages     = {23951--23959},
  year      = {2025},
  doi       = {10.1609/aaai.v39i22.34568},
  url       = {https://ojs.aaai.org/index.php/AAAI/article/view/34568}
}

@inproceedings{liu2023mmsafetybench,
  author    = {Xin Liu and Yichen Zhu and Jindong Gu and Yunshi Lan and Chao Yang and Yu Qiao},
  title     = {{MM-SafetyBench}: A Benchmark for Safety Evaluation of Multimodal Large Language Models},
  booktitle = {Computer Vision -- ECCV 2024},
  series    = {Lecture Notes in Computer Science},
  volume    = {15114},
  pages     = {386--403},
  publisher = {Springer},
  year      = {2025},
  doi       = {10.1007/978-3-031-72992-8_22},
  url       = {https://link.springer.com/chapter/10.1007/978-3-031-72992-8_22}
}

@misc{aiverify2024moonshot,
  author       = {{AI Verify Foundation}},
  title        = {Project {Moonshot}: {Violent Durian} Attack Module},
  year         = {2024},
  howpublished = {Moonshot documentation and source code},
  url          = {https://aiverify-foundation.github.io/moonshot/resources/attack_modules/},
  note         = {Source: \url{https://github.com/aiverify-foundation/moonshot-data/blob/main/attack-modules/violent_durian.py}; accessed 2026-08-31}
}

@misc{pyrit,
  author       = {{Microsoft, AI Red Team}},
  title        = {{PyRIT}: The {Python Risk Identification Tool} for Generative {AI}},
  year         = {2024},
  howpublished = {Software and technical report},
  doi          = {10.48550/arXiv.2410.02828},
  url          = {https://github.com/microsoft/PyRIT},
  note         = {Official repository-recommended citation; released 2024-02-21}
}

@article{garak,
  author  = {Leon Derczynski and Erick Galinkin and Jeffrey Martin and Subho Majumdar and Nanna Inie},
  title   = {garak: A Framework for Security Probing Large Language Models},
  journal = {arXiv preprint arXiv:2406.11036},
  year    = {2024},
  url     = {https://arxiv.org/abs/2406.11036}
}

@misc{deepteam,
  author       = {Jeffrey Ip and Kritin Vongthongsri},
  title        = {deepteam},
  year         = {2026},
  howpublished = {Software, version 1.0.9},
  url          = {https://github.com/confident-ai/deepteam},
  note         = {Released 2026-08-13; Apache-2.0; accessed 2026-08-31}
}

@inproceedings{mazeika2024harmbench,
  author    = {Mantas Mazeika and Long Phan and Xuwang Yin and Andy Zou and Zifan Wang and Norman Mu and Elham Sakhaee and Nathaniel Li and Steven Basart and Bo Li and David Forsyth and Dan Hendrycks},
  title     = {{HarmBench}: A Standardized Evaluation Framework for Automated Red Teaming and Robust Refusal},
  booktitle = {Proceedings of the 41st International Conference on Machine Learning},
  pages     = {35181--35224},
  year      = {2024},
  volume    = {235},
  series    = {Proceedings of Machine Learning Research},
  publisher = {PMLR},
  url       = {https://proceedings.mlr.press/v235/mazeika24a.html}
}

@inproceedings{chao2024jailbreakbench,
  author    = {Patrick Chao and Edoardo Debenedetti and Alexander Robey and Maksym Andriushchenko and Francesco Croce and Vikash Sehwag and Edgar Dobriban and Nicolas Flammarion and George J. Pappas and Florian Tram{\`e}r and Hamed Hassani and Eric Wong},
  title     = {{JailbreakBench}: An Open Robustness Benchmark for Jailbreaking Large Language Models},
  booktitle = {Advances in Neural Information Processing Systems},
  volume    = {37},
  year      = {2024},
  doi       = {10.52202/079017-1745},
  url       = {https://proceedings.neurips.cc/paper_files/paper/2024/hash/63092d79154adebd7305dfd498cbff70-Abstract-Datasets_and_Benchmarks_Track.html}
}

@article{zou2023universal,
  author  = {Andy Zou and Zifan Wang and Nicholas Carlini and Milad Nasr and J. Zico Kolter and Matt Fredrikson},
  title   = {Universal and Transferable Adversarial Attacks on Aligned Language Models},
  journal = {arXiv preprint arXiv:2307.15043},
  year    = {2023},
  url     = {https://arxiv.org/abs/2307.15043}
}

@inproceedings{liu2024autodan,
  author    = {Xiaogeng Liu and Nan Xu and Muhao Chen and Chaowei Xiao},
  title     = {{AutoDAN}: Generating Stealthy Jailbreak Prompts on Aligned Large Language Models},
  booktitle = {The Twelfth International Conference on Learning Representations},
  year      = {2024},
  url       = {https://openreview.net/forum?id=7Jwpw4qKkb}
}

@article{li2023deepinception,
  author  = {Xuan Li and Zhanke Zhou and Jianing Zhu and Jiangchao Yao and Tongliang Liu and Bo Han},
  title   = {{DeepInception}: Hypnotize Large Language Model to Be Jailbreaker},
  journal = {arXiv preprint arXiv:2311.03191},
  year    = {2023},
  url     = {https://arxiv.org/abs/2311.03191}
}

@inproceedings{zheng2023judging,
  author    = {Lianmin Zheng and Wei-Lin Chiang and Ying Sheng and Siyuan Zhuang and Zhanghao Wu and Yonghao Zhuang and Zi Lin and Zhuohan Li and Dacheng Li and Eric P. Xing and Hao Zhang and Joseph E. Gonzalez and Ion Stoica},
  title     = {Judging {LLM-as-a-Judge} with {MT-Bench} and {Chatbot Arena}},
  booktitle = {Advances in Neural Information Processing Systems},
  volume    = {36},
  year      = {2023},
  doi       = {10.52202/075280-2020},
  url       = {https://proceedings.neurips.cc/paper_files/paper/2023/hash/91f18a1287b398d378ef22505bf41832-Abstract-Datasets_and_Benchmarks.html}
}

@inproceedings{souly2024strongreject,
  author    = {Alexandra Souly and Qingyuan Lu and Dillon Bowen and Tu Trinh and Elvis Hsieh and Sana Pandey and Pieter Abbeel and Justin Svegliato and Scott Emmons and Olivia Watkins and Sam Toyer},
  title     = {A {StrongREJECT} for Empty Jailbreaks},
  booktitle = {Advances in Neural Information Processing Systems},
  volume    = {37},
  year      = {2024},
  doi       = {10.52202/079017-3984},
  url       = {https://proceedings.neurips.cc/paper_files/paper/2024/hash/e2e06adf560b0706d3b1ddfca9f29756-Abstract-Datasets_and_Benchmarks_Track.html}
}

@inproceedings{han2024wildguard,
  author    = {Seungju Han and Kavel Rao and Allyson Ettinger and Liwei Jiang and Bill Yuchen Lin and Nathan Lambert and Yejin Choi and Nouha Dziri},
  title     = {{WildGuard}: Open One-stop Moderation Tools for Safety Risks, Jailbreaks, and Refusals of {LLM}s},
  booktitle = {Advances in Neural Information Processing Systems},
  volume    = {37},
  year      = {2024},
  doi       = {10.52202/079017-0261},
  url       = {https://proceedings.neurips.cc/paper_files/paper/2024/hash/0f69b4b96a46f284b726fbd70f74fb3b-Abstract-Datasets_and_Benchmarks_Track.html}
}

@article{openrt2026,
  author  = {Xin Wang and Yunhao Chen and Juncheng Li and Yixu Wang and Yang Yao and Tianle Gu and Jie Li and Yan Teng and Yingchun Wang and Xia Hu},
  title   = {{OpenRT}: An Open-Source Red Teaming Framework for Multimodal {LLM}s},
  journal = {arXiv preprint arXiv:2601.01592},
  year    = {2026},
  url     = {https://arxiv.org/abs/2601.01592}
}

@article{mirage2025,
  author  = {Wenhao You and Bryan Hooi and Yiwei Wang and Youke Wang and Zong Ke and Ming-Hsuan Yang and Zi Huang and Yujun Cai},
  title   = {{MIRAGE}: Multimodal Immersive Reasoning and Guided Exploration for Red-Team Jailbreak Attacks},
  journal = {arXiv preprint arXiv:2503.19134},
  year    = {2025},
  url     = {https://arxiv.org/abs/2503.19134}
}

@article{mjad2026,
  author  = {Badhan Chandra Das and Md Tasnim Jawad and Joaquin Molto and M. Hadi Amini and Yanzhao Wu},
  title   = {Multi-turn Jailbreaking Attack in Multi-Modal Large Language Models},
  journal = {arXiv preprint arXiv:2601.05339},
  year    = {2026},
  url     = {https://arxiv.org/abs/2601.05339}
}

\FloatBarrier
\appendix

\section{Appendix}
\label{sec:appendix}

\subsection{Judge Selection}\label{subsec:judge-ab}
We selected the automated judge through an A/B comparison between GPT-4o and GPT-5.2 on a
stratified sample of 190 multi-turn turns (five harm categories, six targets, five
strategies; identical prompt \texttt{2ed5255}, temperature~0). The judges agree on
83.7\% of items, and all 14 disagreements at the success boundary point one way:
GPT-4o labels \emph{Compliance} where GPT-5.2 assigns a lower category (Compliance
rate 12.1\% vs.\ 4.7\%). Because GPT-4o is more permissive at the Compliance boundary, and is itself
the attacker and one of the targets, we use the more conservative, independent GPT-5.2.

\subsection{Human Validation of Automated Judge}
\label{subsec:human-val}

We validate the final judge (\texttt{gpt-5.2}, prompt \texttt{2ed5255}) against human
annotation on a stratified sample of $n{=}170$ response-level items drawn from
multi-turn runs and submitted for automated judging. Because a random sample
would be dominated by refusals and yield too few of the decision-critical classes, we
over-sample the rare strata so agreement can be estimated on the hard boundary: the 170
comprise 50 \textsc{Compliance}, 50 \textsc{Partial\_Compliance}, 26
\textsc{Indirect\_Refusal}, 12 \textsc{Direct\_Refusal}, 2 \textsc{Non\_Responsive}, and 30
content-policy-blocked items. Each item is the \emph{exact} single-response input submitted to the judge
(\texttt{category=unknown}, no dialogue history); the \texttt{attacker\_blocked} stratum is
empty here. Of the 170, $n{=}140$ have both a valid five-level judge label and a
corresponding human label for judge--human comparison; the remaining 30
content-policy-blocked items lack a valid judge label. Inter-rater reliability and per-class precision/recall/F1 are in
Table~\ref{tab:hv-reliab}.

\paragraph{The hard \textsc{Compliance} boundary is intrinsically subjective.}
Observed judge--human agreement on this boundary is $\kappa_{\mathrm{hard}}{=}0.61$ on the
$n{=}140$ validation sample, while human--human agreement is $\kappa_{\mathrm{hard}}{=}0.56$
on the $n{=}50$ doubly-annotated subset (Table~\ref{tab:hv-reliab}). Because these estimates
use different samples, we report them descriptively rather than treating one as higher. We
therefore treat hard \textsc{Compliance} as a subjective stratum and take the
\textbf{soft boundary} (\textsc{Compliance}$\cup$\textsc{Partial} vs.\ refusal)
as the primary human-validated comparison, where agreement is stable for both judge--human
($\kappa_{\mathrm{soft}}{=}0.57$) and human--human ($\kappa_{\mathrm{soft}}{=}0.69$).

\begin{table}[t]
\centering
\small
\resizebox{\columnwidth}{!}{%
\begin{tabular}{lcc}
\toprule
Boundary & Judge--human & Human--human \\
\midrule
5-level (overall)                 & 0.55 & 0.61 \\
Hard: \textsc{Compliance} vs.\ rest          & 0.61 & 0.56 \\
Soft: \textsc{C}$\cup$\textsc{P} vs.\ refusal & 0.57 & 0.69 \\
\bottomrule
\end{tabular}%
}

\vspace{3pt}
\resizebox{\columnwidth}{!}{%
\begin{tabular}{lccc}
\toprule
Label & P & R & F1 \\
\midrule
Compliance          & 0.58 & 0.94 & 0.72 \\
Partial\_Compliance & 0.66 & 0.72 & 0.69 \\
Indirect\_Refusal   & 0.69 & 0.51 & 0.59 \\
Direct\_Refusal     & 0.92 & 0.85 & 0.88 \\
Non\_Responsive     & 1.00 & 0.13 & 0.24 \\
\bottomrule
\end{tabular}%
}
\caption{Judge validation vs.\ human labels. \textbf{Top:} inter-rater
reliability (Cohen's $\kappa$); judge--human values use the $n{=}140$ validation
sample, human--human values use the $n{=}50$ doubly-annotated subset.
\textbf{Bottom:} per-class judge performance on the $n{=}140$ sample (overall agreement 66.4\%, $\kappa{=}0.55$): a
sensitive but slightly over-eager \textsc{Compliance} detector (R\,0.94, P\,0.58), reliable
on \textsc{Direct\_Refusal} (F1\,0.88).}
\label{tab:hv-reliab}
\end{table}

\paragraph{Population-adjusted agreement.}
Because the validation sample is stratified by the judge's predicted label,
its raw aggregate agreement is not population representative. Reweighting by
the observed prediction-stratum frequencies gives hard-boundary
$\kappa=0.67$ and accuracy $=0.98$, and soft-boundary $\kappa=0.63$ and
accuracy $=0.89$. Compliance precision remains 0.58, indicating that the main
measurement uncertainty is concentrated near the C/PC boundary.

\paragraph{Auxiliary findings.}
(i)~All 30 \texttt{content\_policy} ``refused-to-evaluate'' turns were judged by humans as
genuine refusals (23 \textsc{Non\_Responsive}, 7 \textsc{Direct\_Refusal}); none masked a
hidden compliance. (ii)~Among valid single-turn \texttt{gpt-4o}/image responses, none is
classified as full Compliance. Its 71\% refusal rate excludes \textsc{Non\_Responsive}
outputs by definition and therefore should not be interpreted as a 29\% full-compliance rate.

\subsection{Independent Classifier Cross-Check}
\label{subsec:wildguard}

To test whether the automated judge merely reflects idiosyncrasies of the GPT model
family, we cross-check it against \texttt{WildGuard}~\citep{han2024wildguard}, an
open-weight 7B safety classifier from a different lineage (Mistral-based), run locally.
WildGuard independently labels each (request, response) pair for response harmfulness; we
map its binary verdict against the judge's \textsc{Compliance} label (\textsc{hard}) and
\textsc{Compliance}$\cup$\textsc{Partial} (\textsc{soft}). The request shown to WildGuard is
always the bare harmful goal, never the attacker's disguised framing, so the classifier is
asked whether the response is harmful \emph{with respect to the true objective}.
Results are in Table~\ref{tab:wildguard}.

On the single-turn baseline ($n{=}1175$ direct request--response pairs, an apples-to-apples
setting where the judge and WildGuard see the same input), the two agree on \textbf{98.3\%}
of items with Cohen's $\kappa{=}\textbf{0.65}$ (substantial agreement), and WildGuard raises
almost no false positives (2 of 1175). On this single-turn setting, an independent, differently-trained classifier thus
\emph{substantially corroborates} the judge's compliance grading, directly addressing the
concern that the judge is coupled to the GPT-family attacker.
On multi-turn red-team turns ($n{=}290$, a stratified sample drawn across all 200 goals),
WildGuard's binary harmfulness aligns best with the
judge's \textsc{soft} label ($\kappa{=}0.53$, 78\% agreement); on a 140-item
human-labeled subset it also aligns with human \textsc{soft} labels ($\kappa{=}0.44$). It
flags 59\% of responses as harmful, between the judge's hard-\textsc{Compliance} (35\%) and
soft (69\%) rates. WildGuard counts partial compliances
as harmful, so its threshold sits between our hard and soft boundaries; the items it disputes
are predominantly the reframed (fiction/roleplay) partial compliances that motivate the
gray-zone analysis of Section~\ref{sec:grayzone}.

\begin{table}[t]
\centering
\resizebox{\columnwidth}{!}{%
\begin{tabular}{llcc}
\toprule
Setting & vs. & Agree\% & $\kappa$ \\
\midrule
Single-turn ($n{=}1175$) & judge (hard) & 98.3 & 0.65 \\
Single-turn ($n{=}1175$) & judge (soft) & 97.3 & 0.56 \\
Multi-turn ($n{=}290$)   & judge (hard) & 62.1 & 0.28 \\
Multi-turn ($n{=}290$)   & judge (soft) & 77.9 & 0.53 \\
Multi-turn ($n{=}140$)   & human (soft) & 72.9 & 0.44 \\
\bottomrule
\end{tabular}%
}
\caption{WildGuard (independent open-weight classifier) vs.\ the \texttt{gpt-5.2} judge and
human labels. \textsc{hard}: unsafe~$\Leftrightarrow$~\textsc{Compliance}; \textsc{soft}:
unsafe~$\Leftrightarrow$~\textsc{Compliance}$\cup$\textsc{Partial}.}
\label{tab:wildguard}
\end{table}

\subsection{System Interface}

MUSE exposes five complementary browser views. The \emph{Automated Red Teaming} view drives configurable multi-turn attacks (selecting strategy, goal, target model, per-turn modality rotation, and a max-turn budget), while a \emph{Multimodal Test} view supports single-turn evaluation across text, audio, image, and video, returning the model output alongside the automated safety judgment; the \emph{Statistics Dashboard}, \emph{Interactive Chat}, and \emph{Batch Evaluation} views are introduced in Section~\ref{sec:system}. The full interaction flow for the two attack views is shown in the demo video.

\subsection{Per-Modality Single-Turn Refusal Rates}
\label{subsec:refusal-modality}

Table~\ref{tab:refusal-modality} reports single-turn refusal rates for each
model--modality cell, complementing the same-metric hard/soft/GZW summary in
Table~\ref{tab:baseline}.

\begin{table}[t]
\centering
\resizebox{\columnwidth}{!}{%
\begin{tabular}{@{}l*{5}{c}@{}}
\toprule
\multirow{2}{*}{\textbf{Model}}
  & \multirow{2}{*}{\textbf{Text}}
  & \multirow{2}{*}{\textbf{Image}}
  & \multirow{2}{*}{\textbf{Audio}}
  & \multicolumn{2}{c}{\textbf{Video}} \\
\cmidrule(l){5-6}
  &  &  &  & \textbf{Comb.} & \textbf{Split} \\
\midrule
Claude Sonnet~4.6 & 93\% & 96\% & -- & -- & -- \\
GPT-4o           & 96\% & 71\%$^{\mathrm{a}}$ & -- & -- & -- \\
Gemini 2.5 Flash & 92\% & 94\% & 93\% & 91\% & 90\% \\
Gemini 3 Flash   & 92\% & 96\% & 95\% & 92\% & 93\% \\
Qwen2.5-Omni     & 99\% & 96\% & 99\% & 92\% & 94\% \\
Qwen3-Omni       & 97\% & 97\% & 95\% & 94\% & 98\% \\
\bottomrule
\end{tabular}%
}
\caption{Per-modality single-turn refusal rates (refusal\,=\,Direct\,+\,Indirect Refusal) over all 200 goals (AdvBench-100\,+\,JailbreakBench-100); the content filter blocks a few judge calls, so per-cell denominators are $198$--$200$. \emph{Comb.}\ and \emph{Split} are combined (audio+video) and split (separate-track) video. Claude Sonnet~4.6 and GPT-4o accept only text and image (``--''). $^{\mathrm{a}}$GPT-4o under image: no valid response is classified as full Compliance; Non-Responsive outputs are not included in the refusal numerator (Appendix~\ref{subsec:human-val}).}
\label{tab:refusal-modality}
\end{table}

\subsection{Average Turns to Success}

Table~\ref{tab:avg-turns} reports the mean turns to first Compliance among runs that eventually succeed. This success-conditioned statistic describes the timing of successful trajectories and should not be interpreted as population-level convergence speed.

\begin{table}[H]
\centering
\scriptsize
\setlength{\tabcolsep}{2.5pt}        
\renewcommand{\arraystretch}{0.92}   

\resizebox{\columnwidth}{!}{%
\begin{tabular}{@{}lcccccc@{}}
\toprule
\textbf{Strategy} &
\makecell{\textbf{Claude}\\\textbf{Sonnet~4.6}} &
\makecell{\textbf{GPT-4o}} &
\makecell{\textbf{Gemini}\\\textbf{2.5 Flash}} &
\makecell{\textbf{Gemini}\\\textbf{3 Flash}} &
\makecell{\textbf{Qwen}\\\textbf{2.5-Omni}} &
\makecell{\textbf{Qwen}\\\textbf{3-Omni}} \\
\midrule
Crescendo        & 5.8 & 6.9 & 6.3 & 6.4 & 7.0 & 6.7 \\
ITMS-Crescendo   & \textbf{4.5}\,{\scriptsize($-1.3$)} & \textbf{6.4}\,{\scriptsize($-0.5$)} & \textbf{6.0}\,{\scriptsize($-0.3$)} & \textbf{5.9}\,{\scriptsize($-0.5$)} & \textbf{6.6}\,{\scriptsize($-0.4$)} & \textbf{6.5}\,{\scriptsize($-0.2$)} \\
\addlinespace
Violent Durian   & -- & 4.7 & 4.5 & 3.8 & 4.5 & 4.8 \\
ITMS-VD          & -- & 5.2\,{\scriptsize($+0.5$)} & \textbf{4.2}\,{\scriptsize($-0.3$)} & 4.5\,{\scriptsize($+0.7$)} & \textbf{4.2}\,{\scriptsize($-0.3$)} & 5.2\,{\scriptsize($+0.4$)} \\
\bottomrule
\end{tabular}%
}

\caption{Mean turns to first Compliance among successful runs; all conditions were run on 200 goals. Parenthesized $\Delta$ values are relative to the base strategy; \textbf{bold} = fewer turns among successful runs. ``--'' marks cells with no hard-Compliance successes from which to estimate a mean; Claude Sonnet~4.6 produces no hard-Compliance successes under either Violent Durian variant. Among successful runs, ITMS-Crescendo reaches the first Compliance in fewer turns for all 6 of 6 models.}
\label{tab:avg-turns}
\end{table}

\subsection{Five-Level Response Distribution}
\label{subsec:five-level-dist}

Table~\ref{tab:five-level-dist} gives the run-level five-category
\emph{severity-ceiling} distribution by strategy, aggregated over all six
targets on the 200-goal Table~\ref{tab:asr-matrix} scope. Each valid run is
assigned the single most-engaged label it attains across its turns, ordered
\textsc{Compliance} $>$ \textsc{Partial\_Compliance} $>$
\textsc{Indirect\_Refusal} $>$ \textsc{Direct\_Refusal} $>$
\textsc{Non\_Responsive}.
The \textsc{Compliance} column therefore corresponds to hard ASR over the
same run pool as Table~\ref{tab:asr-matrix}, and
\textsc{Compliance}$+$\textsc{Partial\_Compliance} to soft ASR. The strategies occupy visibly different regions of the
taxonomy: PAIR is dominated by \textsc{Compliance}, both Crescendo variants by
\textsc{Partial\_Compliance} and \textsc{Indirect\_Refusal}, and both Violent
Durian variants retain a substantial \textsc{Direct\_Refusal} share. Collapsing
to a binary label would merge these qualitatively different profiles.

\begin{table}[t]
\centering
\setlength{\tabcolsep}{4pt}
\resizebox{\columnwidth}{!}{%
\begin{tabular}{@{}lccccc@{}}
\toprule
\textbf{Strategy} & \textbf{C} & \textbf{PC} & \textbf{IR} & \textbf{DR} & \textbf{NR} \\
\midrule
Crescendo      & 28.5 & 44.1 & 27.5 & 0.0  & 0.0 \\
ITMS-Crescendo & 24.0 & 43.9 & 32.1 & 0.0  & 0.0 \\
Violent Durian & 27.5 & 8.7  & 41.6 & 22.1 & 0.1 \\
ITMS-VD        & 24.1 & 10.5 & 43.2 & 22.2 & 0.1 \\
PAIR           & 74.8 & 18.1 & 6.0  & 0.8  & 0.3 \\
\bottomrule
\end{tabular}%
}
\caption{Run-level five-category severity-ceiling distribution by strategy
(\% of valid runs; ${\approx}1{,}190$ runs per strategy,
Table~\ref{tab:asr-matrix} scope; $R_e$ excluded). Each run is assigned its
most-engaged label across turns
(\textsc{C}$>$\textsc{PC}$>$\textsc{IR}$>$\textsc{DR}$>$\textsc{NR}).
\textsc{C}=Compliance, \textsc{PC}=Partial Compliance,
\textsc{IR}=Indirect Refusal, \textsc{DR}=Direct Refusal, \textsc{NR}=Non-Responsive.
Rows sum to 100\% up to rounding. \textsc{C} corresponds to hard ASR and
\textsc{C}$+$\textsc{PC} to soft ASR over the same run pool.}
\label{tab:five-level-dist}
\end{table}

\end{document}